\documentclass{llncs}
\usepackage{amsfonts}
\usepackage{amssymb}
\usepackage{amsmath,amssymb,amscd,amsfonts,rotating}
\usepackage{graphicx}
\usepackage{epsfig,epstopdf}
\usepackage{multirow}
\usepackage{booktabs}
\usepackage{url}
\usepackage{xcolor,colortbl,subfigure,wrapfig}
\usepackage{tikz}
\usepackage{makeidx}  
\usetikzlibrary{arrows}
\usetikzlibrary{fit}
\usepackage{algorithm}
\usepackage{algpseudocode}

\newcommand{\bs}{\boldsymbol}
\newcommand{\sigmoid}{\text{sigmoid}}

\newcommand{\tanhh}{\text{tanh}}

\newcommand{\tstart}{\text{start}}
\newcommand{\tend}{\text{end}}

\newcommand{\ft}{\texttt}
\newcommand{\fmnn}{\text{FNN}}
\newcommand{\snn}{\text{SNN}}
\newcommand{\snnrbm}{\text{SNN-RBM}}
\newcommand{\snndae}{\text{SNN-DAE}}

\begin{document}
\title{Deep Learning over Multi-field Categorical Data}
\subtitle{-- A Case Study on User Response Prediction}

\author{Weinan Zhang\inst{1} \and Tianming Du\inst{1,2} \and Jun Wang\inst{1}}
\institute{$^1$University College London, London, United Kingdom\\
$^2$RayCloud Inc., Hangzhou, China\\
\email{$^1$\{w.zhang,j.wang\}@cs.ucl.ac.uk, $^2$dutianming@quicloud.cn}}

\maketitle

\vspace{-10pt}
\begin{abstract}
Predicting user responses, such as click-through rate and conversion rate, are critical in many web applications including web search, personalised recommendation, and online advertising. Different from continuous raw features that we usually found in the image and audio domains, the input features in web space are always of multi-field and are mostly discrete and categorical while their dependencies are little known. Major user response prediction models have to either limit themselves to linear models or require manually building up high-order combination features. The former loses the ability of exploring feature interactions, while the latter results in a heavy computation in the large feature space. To tackle the issue, we propose two novel models using deep neural networks (DNNs) to automatically learn effective patterns from categorical feature interactions and make predictions of users' ad clicks. To get our DNNs efficiently work, we propose to leverage three feature transformation methods, i.e., factorisation machines (FMs), restricted Boltzmann machines (RBMs) and denoising auto-encoders (DAEs). This paper presents the structure of our models and their efficient training algorithms. The large-scale experiments with real-world data demonstrate that our methods work better than major state-of-the-art models.
\end{abstract}

\section{Introduction}
User response (e.g., click-through or conversion) prediction plays a critical part in many web applications including web search, recommender systems, sponsored search, and display advertising. In online advertising, for instance, the ability of targeting individual users is the key advantage compared to traditional offline advertising.  All these targeting techniques, essentially,  rely on the system function of predicting whether a specific user will think the potential ad is ``relevant", i.e., the probability that the user in a certain context will click a given ad \cite{broder2008computational}. Sponsored search, contextual advertising, and the recently emerged real-time bidding (RTB) display advertising all heavily rely on the ability of learned models to predict ad click-through rates (CTR) \cite{richardson2007predicting,zhang2014optimal}. The applied CTR estimation models today are mostly linear, ranging from logistic regression \cite{richardson2007predicting} and naive Bayes \cite{hand2001idiot} to FTRL logistic regression \cite{mcmahan2013ad} and Bayesian probit regression \cite{graepel2010web}, all of which are based on a huge number of sparse features with one-hot encoding \cite{beck2000high}. Linear models have advantages of easy implementation, efficient learning but relative low performance because of the failure of learning the non-trivial patterns to catch the interactions between the assumed (conditionally) independent raw features \cite{graepel2010web}.  Non-linear models, on the other hand, are  able to utilise different feature combinations and thus could potentially improve estimation performance. For example, factorisation machines (FMs) \cite{oentaryo2014predicting} map the user and item binary features into a low dimensional continuous space. And the feature interaction is automatically explored via vector inner \text{product}. Gradient boosting trees \cite{trofimov2012using} automatically learn feature combinations while growing each decision/regression tree. However, these models cannot make use of all possible combinations of different features \cite{juan2014idiot}. In addition, many models require feature engineering that manually designs what the inputs should be. Another problem of the mainstream ad CTR estimation models is that most prediction models have shallow structures and have limited expression to model the underlying patterns from complex and massive data \cite{he2014practical}. As a result, their data modelling and generalisation ability is still restricted.

Deep learning \cite{lecun2015deep} has become successful in computer vision \cite{krizhevsky2012imagenet},  speech recognition \cite{graves2013speech}, and natural language processing (NLP) \cite{huang2013learning,shen2014latent} during \text{recent} five years. As visual, aural, and textual signals are known to be spatially and/or temporally correlated, the newly introduced unsupervised training on deep structures \cite{hinton2006reducing} would be able to explore such \emph{local} dependency and establish a \emph{dense} representation of the feature space, making neural network models effective in learning high-order features directly from the raw feature input.  With such learning ability, deep learning would be a good candidate to estimate online user response rate such as ad CTR. However, most input features in CTR estimation are of multi-field and are discrete categorical features, e.g., the user location city (\ft{London}, \ft{Paris}), device type (\ft{PC}, \ft{Mobile}), ad category (\ft{Sports}, \ft{Electronics}) etc., and their local dependencies (thus the sparsity in the feature space) are unknown.  Therefore, it is of great interest to see how deep learning improves the CTR estimation via learning feature representation on such large-scale multi-field discrete categorical features. To our best knowledge, there is no previous literature of ad CTR estimation using deep learning methods thus far\footnote{Although the leverage of deep learning models on ad CTR estimation has been claimed in industry (e.g., \cite{zou2014mariana}), there is no detail of the models or implementation.}. In addition, training deep neural networks (DNNs) on a large input feature space requires tuning a huge number of parameters, which is computationally expensive. For instance, unlike image and audio cases, we have about 1 million binary input features and 100 hidden units in the first layer; then it requires 100 million links to build the first layer neural network.

In this paper, we take ad CTR estimation as a working example to study deep learning over a large multi-field categorical feature space by using embedding methods in both supervised and unsupervised fashions. We introduce two types of deep learning models, called Factorisation Machine supported Neural Network (\fmnn) and Sampling-based Neural Network (\snn). Specifically, \fmnn~with a supervised-learning embedding layer using factorisation machines \cite{rendle2012factorization} is proposed to efficiently reduce the dimension from sparse features to dense continuous features. The second model \snn~is a deep neural network powered by a sampling-based restricted Boltzmann machine (\snnrbm) or a sampling-based denoising auto-encoder (\snndae) with a proposed negative sampling method. Based on the embedding layer, we build multiple layers neural nets with full connections to explore non-trivial data patterns. Our experiments on multiple real-world advertisers' ad click data have demonstrated the consistent improvement of CTR estimation from our proposed models over the state-of-the-art ones.



\vspace{-5pt}
\section{Related Work}
\vspace{-5pt}


Click-through rate, defined as the probability of the ad click from a specific user on a displayed ad, is essential in online advertising \cite{wang2010click}. In order to maximise revenue and user satisfaction, online advertising platforms must predict the expected user behaviour for each displayed ad and maximise the expectation that users will click. The majority of current models use logistic regression based on a set of sparse binary features converted from the original categorical features via one-hot encoding \cite{lee2012estimating,richardson2007predicting}. Heavy engineering efforts are needed to design features such as locations, top unigrams, combination features, etc. \cite{he2014practical}.

Embedding very large feature vector into low-dimensional vector spaces is useful for prediction task as it reduces the data and model complexity and improves both the effectiveness and the efficiency of the training and prediction. Various methods of embedding architectures have been proposed \cite{tang2015line,kurashima2014probabilistic}.
Factorisation machine (FM) \cite{rendle2012factorization}, originally proposed for collaborative filtering recommendation, is regarded as one of the most successful embedding models. FM naturally has the capability of estimating interactions between any two features via mapping them into vectors in a low-rank latent space.


Deep Learning \cite{bengio2009learning} is a branch of artificial intelligence research that attempts to develop the techniques that will allow computers to handle complex tasks such as recognition and prediction at high performance. Deep neural \text{networks} (DNNs) are able to extract the hidden structures and intrinsic patterns at different levels of abstractions from training data. DNNs have been successfully applied in computer vision \cite{zeiler2011adaptive}, speech recognition \cite{deng2013deep} and natural language processing (NLP) \cite{collobert2011natural,huang2013learning,shen2014latent}. Furthermore, with the help of unsupervised pre-training, we can get good feature representation which guides the learning towards \text{basins} of attraction of minima that support better generalisation from the training data \cite{erhan2010does}. Usually, these deep models have two stages in learning \cite{hinton2006reducing}: the first stage performs model initialisation via unsupervised learning (i.e., the restricted Boltzmann machine or stacked denoising auto-encoders) to make the model catch the input data distribution; the second stage involves a fine tuning of the initialised model via supervised learning with back-propagation. The novelty of our deep learning models lies in the first layer initialisation, where the input raw features are high dimensional and sparse binary features converted from the raw categorical features, which makes it hard to train traditional DNNs in large scale. Compared with the word-embedding techniques used in NLP \cite{huang2013learning,shen2014latent}, our models deal with more general multi-field categorical features without any assumed data structures such as word alignment and letter-n-gram etc.

\section{DNNs for CTR Estimation given Categorical Features}

\begin{figure}[t]
  \centering
  \vspace{-30pt}
  \includegraphics[width=0.75\columnwidth]{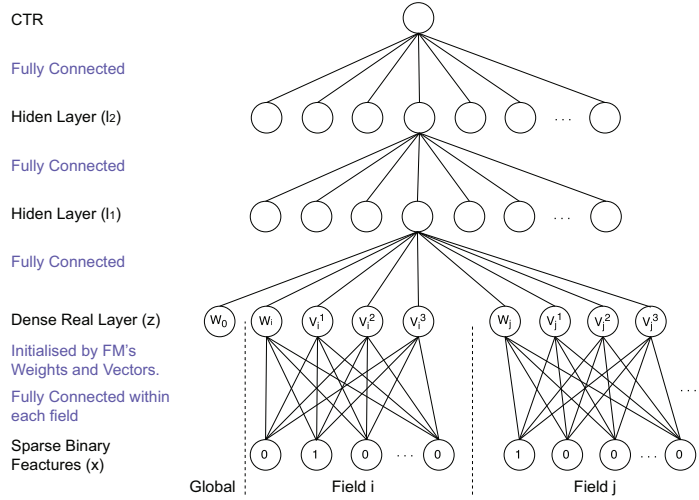}\\
  \caption{A 4-layer \fmnn~model structure.}\label{fig:modelone}
\end{figure}

In this section, we discuss the two proposed DNN architectures in detail, namely Factorisation-machine supported Neural Networks (\fmnn) and Sampling-based Neural Networks (\snn).
The input categorical features are field-wise one-hot encoded. For each field, e.g., \ft{city}, there are multiple units, each of which represents a specific value of this field, e.g., \ft{city=London}, and there is only one positive (1) unit, while all others are negative (0). The encoded features, denoted as $\bs{x}$, are the input of many CTR estimation models \cite{richardson2007predicting,lee2012estimating} as well as our DNN models, as depicted at the bottom layer of Figure~\ref{fig:modelone}.

\subsection{Factorisation-machine supported Neural Networks (\fmnn)}\label{sec:fmnn}
Our first model \fmnn~is based on the factorisation machine as the bottom layer. The network structure is shown in Figure~\ref{fig:modelone}. With a top-down description, the output unit is a real number $\hat{y} \in (0,1)$ as predicted CTR, i.e., the probability of a specific user clicking a given ad in a certain context:
\begin{align}
\hat{y} =\sigmoid(\bs{W}_{3}\bs{l}_{2}+b_{3}),   \label{eq:xdef1}
\end{align}
where $\sigmoid (x)=1/(1+e^{-x})$ is the logistic activation function, $\bs{W}_{3}\in\mathbb{R}^{1 \times L}$, $b_{3}\in \mathbb{R}$  and $\bs{l}_{2} \in \mathbb{R}^{L}$ as input for this layer. The calculation of $\bs{l}_{2}$ is
\begin{align}
\bs{l}_{2}=\tanhh(\bs{W}_{2}\bs{l}_{1}+\bs{b}_{2}),   \label{eq:xdef2}
\end{align}
where $\tanh(x)=(1-e^{-2x})/(1+e^{-2x})$, $\bs{W}_{2}\in\mathbb{R}^{L\times M}$, $\bs{b}_{2}\in \mathbb{R}^{L}$ and $\bs{l}_{1} \in \mathbb{R}^{M}$. We choose $\tanh(\cdot)$ as it has optimal empirical learning performance than other activation functions, as will be discussed in Section~\ref{sec:hyper-tuning}. Similarly,
\begin{align}
\bs{l}_{1}=\tanhh(\bs{W}_{1}\bs{z}+\bs{b}_{1}),   \label{eq:xdef3}
\end{align}
where $\bs{W}_{1}\in\mathbb{R}^{M\times J}$, $\bs{b}_{1}\in \mathbb{R}^{M}$ and $\bs{z} \in \mathbb{R}^{J}$.
\begin{align}
\bs{z}=(w_{0}, \bs{z}_{1}, \bs{z}_{2}, ...\bs{z}_{i}, ..., \bs{z}_{n}),
 \label{eq:xdef4}
\end{align}
where $w_{0} \in \mathbb{R}$ is a global scalar parameter and $n$ is the number of fields in total. $\bs{z}_{i} \in\mathbb{R}^{K+1}$ is a parameter vectors for the $i$-th field in factorisation machines:
\begin{equation}
\bs{z}_{i}=\bs{W}_{0}^{i}\cdot \bs{x}[\tstart_{i}:\tend_{i}] = (w_i, v_i^1, v_i^2, \ldots, v_i^K),
 \label{eq:xdef5}
\end{equation}
where $\tstart_{i}$ and $\tend_{i}$ are starting and ending feature indexes of the $i$-th field, $\bs{W}_{0}^{i} \in \mathbb{R}^{(K+1) \times (\tend_{i}-\tstart_{i}+1)} $ and $\bs{x}$ is the input vector as described at beginning. All weights $\bs{W}_{0}^{i}$ are initialised with the bias term $w_{i}$ and vector $\bs{v}_{i}$ respectively (e.g., $\bs{W}_{0}^{i}[0]$ is initialised by $w_{i}$, $\bs{W}_{0}^{i}[1]$ is initialised by $v_{i}^{1}$, $\bs{W}_{0}^{i}[2]$ is initialised by $v_{i}^{2}$, etc.). In this way, $\bs{z}$ vector of the first layer is initialised as shown in Figure~\ref{fig:modelone} via training a factorisation machine (FM) \cite{rendle2012factorization}:
\begin{equation}
 y_{\text{FM}}(\bs{x}):=\sigmoid \Big( w_{0}+\sum_{i=1}^{N} w_{i} x_{i} + \sum_{i=1}^{N}\sum_{j=i+1}^{N}\langle\bs{v}_{i},\bs{v}_{j}\rangle x_{i} x_{j} \Big), \label{eq:xdef}
\end{equation}
where each feature $i$ is assigned with a bias weight $w_i$ and a $K$-dimensional vector $\bs{v}_i$ and the feature interaction is modelled as their vectors' inner product $\langle\bs{v}_{i},\bs{v}_{j}\rangle$. In this way, the above neural nets can learn more efficiently from factorisation machine representation so that the computational complexity problem of the high-dimensional binary inputs has been naturally bypassed. Different hidden layers can be regarded as different internal functions capturing different forms of representations of the data instance. For this reason, this model has more abilities of catching intrinsic data patterns and leads to better performance.

The idea using FM in the bottom layer is ignited by Convolutional Neural Networks (CNNs) \cite{fukushima1980neocognitron}, which exploit spatially local correlation by enforcing a local connectivity pattern between neurons of adjacent layers.  Similarly, the inputs of hidden layer 1 are connected to the input units of a specific field. Also, the bottom layer is not fully connected as FM performs a field-wise training for one-hot sparse encoded input, allowing local sparsity, illustrated as the dash lines in Figure~\ref{fig:modelone}.  FM learns good structural data representation in the latent space, helpful for any further model to build on.  A subtle difference, though, appears between the product rule of FM and the sum rule of DNN for combination. However, according to \cite{kittler1998combining}, if the observational discriminatory information is highly ambiguous (which is true in our case for ad click behaviour), the posterior weights (from DNN) will not deviate dramatically from the prior (FM).

Furthermore, the weights in hidden layers (except the FM layer) are initialised by layer-wise RBM pre-training \cite{bengio2007greedy} using contrastive divergence \cite{hinton2002training}, which effectively preserves the information in input dataset as detailed in \cite{hinton2006reducing,hinton2010practical}.  The initial weights for FMs are trained by stochastic gradient descent (SGD), as detailed in \cite{rendle2012factorization}. Note that we only need to update weights which connect to the positive input units, which largely reduces the computational complexity. After pre-training of the FM and upper layers, supervised fine-tuning (back propagation) is applied to minimise loss function of cross entropy:
\begin{align}
L(y,\hat{y})= -y\log \hat{y} - (1-y)\log(1-\hat{y})  \label{eq:xdefloss},
\end{align}
where $\hat{y}$ is the predicted CTR in Eq.~(\ref{eq:xdef1}) and $y$ is the binary click ground-truth label. Using the chain rule of back propagation, the \fmnn~weights including FM weights can be efficiently updated. For example, we update FM layer weights via
\begin{align}
\frac{\partial L(y,\hat{y})}{\partial \bs{W}_{0}^{i}} &=  \frac{\partial L(y,\hat{y})}{\partial \bs{z}_i} \frac{\partial \bs{z}_i}{\partial \bs{W}_{0}^{i}} =\frac{\partial L(y,\hat{y})}{\partial \bs{z}_i} \bs{x}[\text{start}_i:\text{end}_i] \\
\bs{W}_{0}^{i} &\leftarrow \bs{W}_{0}^{i}-\eta \cdot \frac{\partial L(y,\hat{y})}{\partial \bs{z}_i} \bs{x}[\text{start}_i:\text{end}_i].  \label{eq:def_grad}
\end{align}
Due to the fact that the majority entries of $\bs{x}[\text{start}_i:\text{end}_i]$ are 0, we can accelerate fine-tuning by updating weights linking to positive units only.

%

\begin{figure}[t]
  \centering
  \vspace{-35pt}
  \includegraphics[width=1.1\columnwidth]{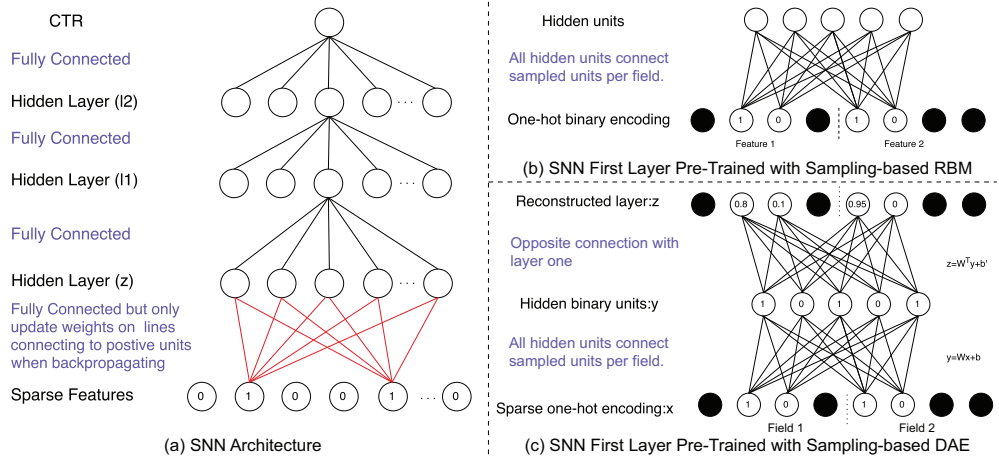}\\
  \caption{A 4-layer \snn~architecture and two first-layer pre-training methods.}\label{fig:snn}
\end{figure}

\subsection{Sampling-based Neural Networks (\snn)}\label{sec:snn}
The structure of the second model \snn~is shown in Figure~\ref{fig:snn}(a). The difference between \snn~and \fmnn~lies in the structure and training method in the bottom layer. \snn's bottom layer is fully connected with sigmoid activation function:
\begin{equation}
\bs{z}=\sigmoid(\bs{W}_{0}\bs{x}+\bs{b}_{0}).   \label{eq:defmodtwo}
\end{equation}

To initialise the weights of the bottom layer, we tried both restricted Boltzmann machine (RBM) \cite{hinton2010practical} and denoising auto-encoder (DAE) \cite{bengio2013generalized} in the pre-training stage. In order to deal with the computational problem of training large sparse one-hot encoding data, we propose a sampling-based RBM (Figure~\ref{fig:snn}(b), denoted as \snnrbm) and a sampling-based DAE in (Figure~\ref{fig:snn}(c), denoted as \snndae) to efficiently calculate the initial weights of the bottom layer.

Instead of modelling the whole feature set for each training instance set, for each feature field, e.g., \ft{city}, there is only one positive value feature for each training instance, e.g., \ft{city=London}, we sample $m$ negative units, e.g., \ft{city=Paris} when $m=1$, randomly with value 0. Black units in Figure~\ref{fig:snn}(b) and \ref{fig:snn}(c) are unsampled and thus ignored when pre-training the data instance. With the sampled units, we can train an RBM via contrastive divergence \cite{hinton2002training} and a DAE via SGD with unsupervised approaches to largely reduce the data dimension with high recovery performance. The real-value dense vector is used as the input of the further layers in SNN.

In this way, computational complexity can be dramatically reduced and, in turn, initial weights can be calculated quickly and back-propagation is then performed to fine-tune \snn~model.

\subsection{Regularisation}\vspace{-2pt}
To prevent overfitting, the widely used L2 regularisation term is added to the loss function. For example, the L2 regularisation for \fmnn~in Figure~\ref{fig:modelone} is
\begin{align}
\Omega(\bs{w})=||\bs{W}_{0}||_2^{2}+\sum_{l=1}^3 \Big( ||\bs{W}_{l}||_2^{2}+||\bs{b}_{l}||_2^{2} \Big). \label{eq:xdefreg}
\end{align}
On the other hand, \emph{dropout} \cite{srivastava2014dropout} is a technique which becomes a popular and effective regularisation technique for deep learning during the recent years. We also implement this regularisation and compare them in our experiment.

\vspace{-5pt}
\section{Experiment}\label{sec:exp}
\vspace{-3pt}
\subsection{Experiment Setup}
\textbf{Data.} We evaluate our models based on iPinYou dataset \cite{liao2014ipinyou}, a public real-world display ad dataset with each ad display information and corresponding user click feedback. The data logs are organised by different advertisers and in a row-per-record format. There are 19.50M data instances with 14.79K positive label (click) in total. The features for each data instance are all categorical. Feature examples in the ad log data are \ft{user agent}, partially masked \ft{IP}, \ft{region}, \ft{city}, \ft{ad exchange}, \ft{domain}, \ft{URL}, \ft{ad slot ID}, \ft{ad slot visibility}, \ft{ad slot size}, \ft{ad slot format}, \ft{creative ID}, \ft{user tags}, etc. After one-hot encoding, the number of binary features is 937.67K in the whole dataset. We feed each compared model with these binary-feature data instances and the user click (1) and non-click (0) feedback as the ground-truth labels. In our experiments, we use training data from advertiser 1458, 2259, 2261, 2997, 3386 and the whole dataset, respectively.


\noindent \textbf{Models.} We compare the performance of the following CTR estimation models:
\begin{itemize}
  \item [LR:] Logistic Regression \cite{richardson2007predicting} is a linear model with simple implementation and fast training speed, which is widely used in online advertising estimation.
  \item [FM:] Factorisation Machine \cite{rendle2012factorization} is a non-linear model able to estimate feature interactions even in problems with huge sparsity.
  \item [\fmnn:] Factorisation-machine supported Neural Network is our proposed model as described in Section~\ref{sec:fmnn}.
  \item [\snn:] Sampling-based Neural Network is also our proposed model with sampling-based RBM and DAE pre-training methods for the first layer in Section~\ref{sec:snn}, denoted as \snnrbm~and \snndae~respectively.
\end{itemize}
Our experiment code\footnote{The source code with demo data: \url{https://github.com/wnzhang/deep-ctr}} of both \fmnn~and \snn~is implemented with \ft{Theano}\footnote{\ft{Theano}: \url{http://deeplearning.net/software/theano/}}.

\noindent \textbf{Metric.} To measure the CTR estimation performance of each model, we employ the area under ROC curve (AUC)\footnote{Besides AUC, root mean square error (RMSE) is also tested. However, positive/negative examples are largly unbalanced in ad click scenario, and the empirically best regression model usually provides the predicted CTR close to 0, which results in very small RMSE values and thus the improvement is not well captured.}. The AUC \cite{graepel2010web} metric is a widely used measure for evaluating the CTR performance.

\subsection{Performance Comparison}

\begin{table}[t]
\centering
\vspace{-30pt}
\caption{Overall CTR estimation AUC performance.}\label{tab:performance}
\begin{tabular}{c|ccccc}
& ~~~~LR~~~~ & ~~~~FM~~~~ & ~~~\fmnn~~~ & \snndae & \snnrbm\\ \hline
1458 & 70.42\% & 70.21\% & \textbf{70.52\%} & 70.46\% & 70.49\% \\
2259 & 69.66\% & 69.73\% & \textbf{69.74\%} & 68.08\% & 68.34\% \\
2261 & 62.03\% & 60.97\% & 62.99\% & \textbf{63.72\%} & \textbf{63.72\%} \\
2997 & 60.77\% & 60.87\% & 61.41\% & \textbf{61.58\%} & 61.45\% \\
3386 & 80.30\% & 79.05\% & \textbf{80.56\%} & 79.62\% & 80.07\% \\
all  & 68.81\% & 68.18\% & \textbf{70.70\%} & 69.15\% & 69.15\% \\
\end{tabular}
\end{table}

Table~\ref{tab:performance} shows the results that compare LR, FM, \fmnn~and \snn~with RBM and DAE on 5 different advertisers and the whole dataset. We observe that FM is not significantly better than LR, which means 2-order combination features might not be good enough to catch the underlying data patterns. The AUC performance of the proposed \fmnn~and \snn~is better than the performance of LR and FM on all tested datasets. Based on the latent structure learned by FM, \fmnn~further learns effective patterns between these latent features and provides a consistent improvement over FM. The performance of \snndae~and \snnrbm~is generally consistent, i.e., the relative order of the results of the \snn~are almost the same.

\subsection{Hyperparameter Tuning}\label{sec:hyper-tuning}

Due to the fact that deep neural networks involve many implementation details and need to tune a fairly large number of hyper-parameters, following details show how we implement our models and tune hyperparameters in the models.

We use stochastic gradient descent to learn most of our parameters for all proposed models. Regarding selecting the number of training epochs, we use  early stopping  \cite{prechelt1998automatic}, i.e., the training stops when the validation error increases. We try different learning rate from 1, 0.1, 0.01, 0.001 to 0.0001 and choose the one with optimal performance on the validation dataset.

For negative unit sampling of \snnrbm~and \snndae, we try the negative sample number $m=1,2$ and $4$ per field as described in Section~\ref{sec:snn}, and find $m=2$ produces the best results in most situations. For the activation functions in both models on the hidden layers (as Eqs.~(\ref{eq:xdef3}) and (\ref{eq:xdef2})), we try linear function, sigmoid function and tanh function, and find the result of tanh function is optimal. This might be because the hyperbolic tangent often converges faster than the sigmoid function.



\subsection{Architecture Selection}\label{sec:struct}
In our models, we investigate architectures with 3, 4 and 5 hidden layers by fixing all layer sizes and find the architecture with 3 hidden layers (i.e., 5 layers in total) is the best in terms of AUC performance. However, the range of choosing their layer sizes is exponential in the number of hidden layers. Suppose there is a deep neural network with $L$ hidden layers and each of the hidden layers is trained with a range of hidden units from 100 to 500 with increments of 100, thus there are $5^{L}$ models in total to compare.

\begin{figure}[t]
    \centering
    \vspace{-30pt}
    \subfigure[Architecture]{\includegraphics[width=0.22\columnwidth]{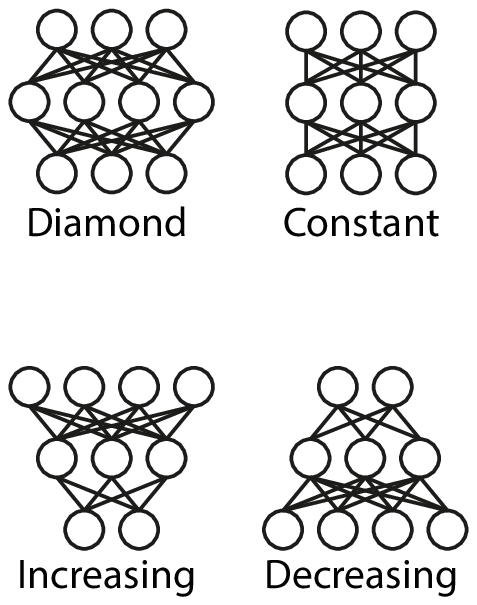}\label{fig:dnn-struct}}
    \subfigure[\snnrbm~on 2997]{\includegraphics[width=0.37\columnwidth]{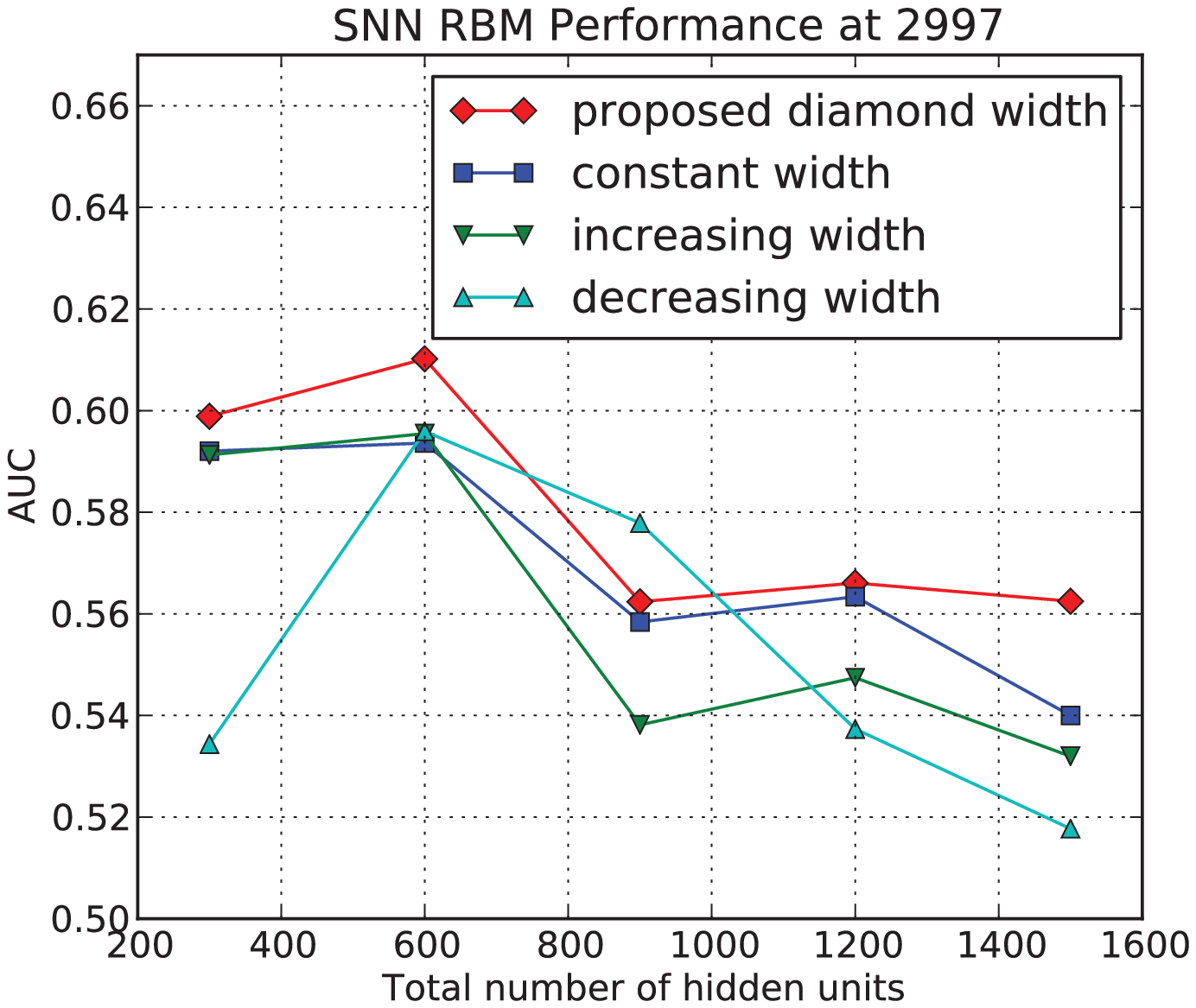}\label{fig:layersize-2997} } %
    \subfigure[\snnrbm~on 3386]{\includegraphics[width=0.37\columnwidth]{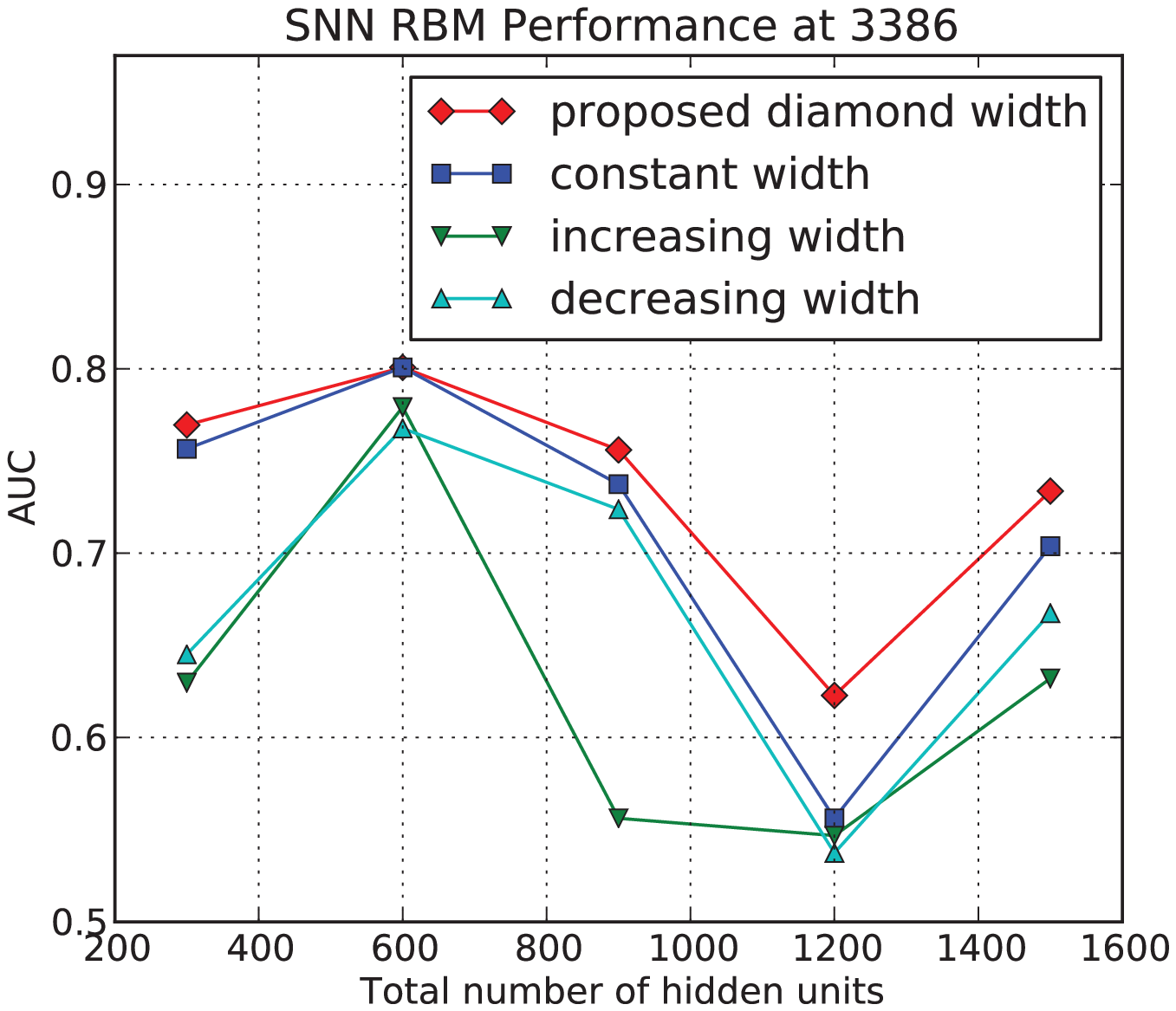}\label{fig:layersize-3386} } %
    \vspace{-10pt} \caption{AUC Performance with different architectures.}\vspace{-10pt}
    \label{fig:layersize}%
\end{figure}

Instead of trying all combinations of hidden units, in our experiment we use another strategy by starting tuning the different hidden layer sizes with the same number of hidden units in all three hidden layers\footnote{Some advanced Bayesian methods for hyperparameter tuning \cite{snoek2012practical} are not considered in this paper and may be investigated in the future work.} since the architecture with equal-size hidden layers is empirically better than the architecture with increasing width or decreasing width in \cite{larochelle2009exploring}. For this reason, we start tuning layer sizes with equal hidden layer sizes. In fact, apart from increasing, constant, decreasing layer sizes, there is a more effective structure, which is the diamond shape of neural networks, as shown in Figure~\ref{fig:dnn-struct}. We compare our diamond shape network with other three shapes of networks and tune the total number of total hidden units on two different datasets shown in Figures~\ref{fig:layersize-2997} and \ref{fig:layersize-3386}. The diamond shape architecture outperforms others in almost all layer size settings. The reason why this diamond shape works might be because this special shape of neural network has certain constraint to the capacity of the neural network, which provides better generalisation on test sets. On the other hand, the performance of diamond architecture picks at the total hidden unit size of 600, i.e., the combination of (200, 300, 100). This depends on the training data observation numbers. Too many hidden units against a limited dataset could cause overfitting.




\vspace{-0pt}
\subsection{Regularisation Comparison}\label{sec:reg}
Neural network training algorithms are very sensitive to the overfitting problem since deep networks have multiple non-linear layers, which makes them very expressive models that can learn very complicated functions. For DNN models, we compared L2 regularisation (Eq.~(\ref{eq:xdefreg})) and dropout \cite{srivastava2014dropout} for preventing complex co-adaptations on the training data. The dropout rate implemented in this experiment refers to the probability of each unit being active.

\begin{figure}[t]
    \centering
    \vspace{-30pt}
    \subfigure[Dropout vs. L2]{{\includegraphics[width=0.33\columnwidth]{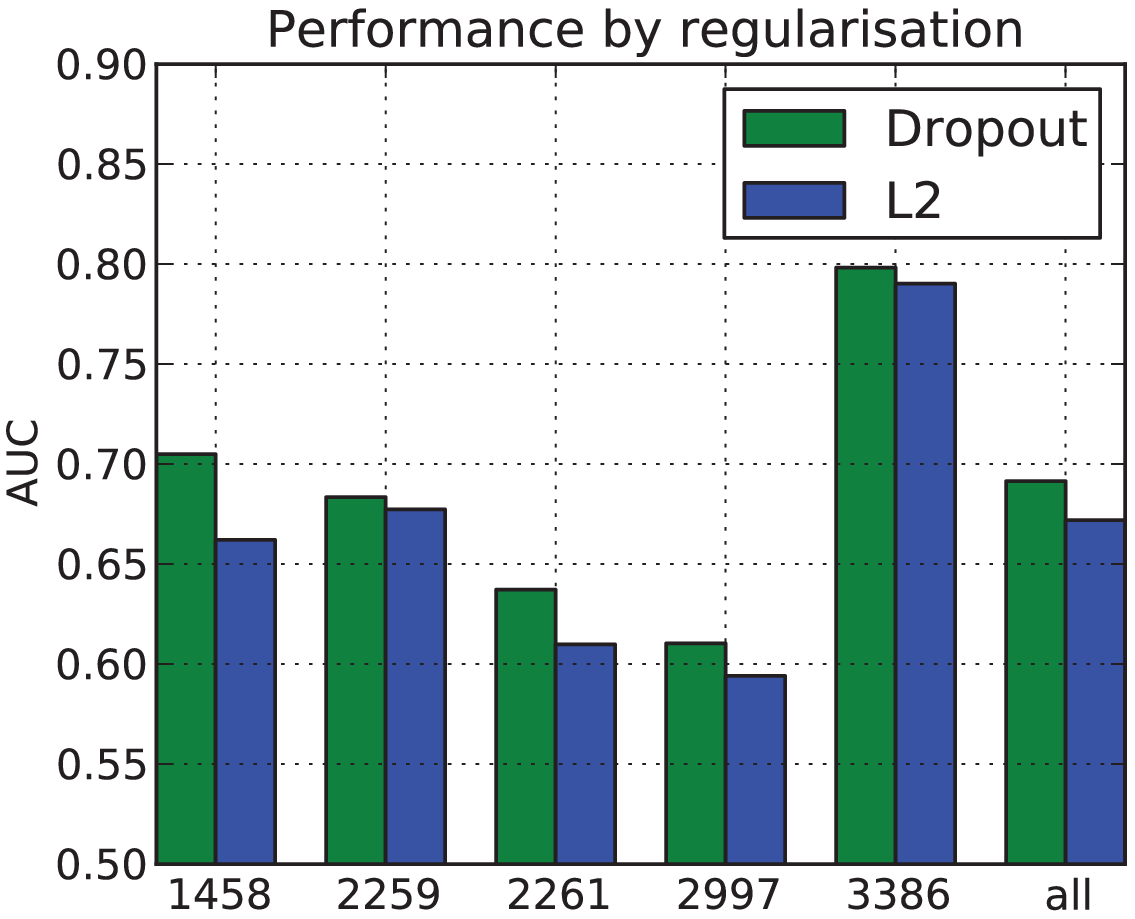} \label{fig:regular}}}%
    \subfigure[\fmnn~on 2997 dataset]{{\includegraphics[width=0.33\columnwidth]{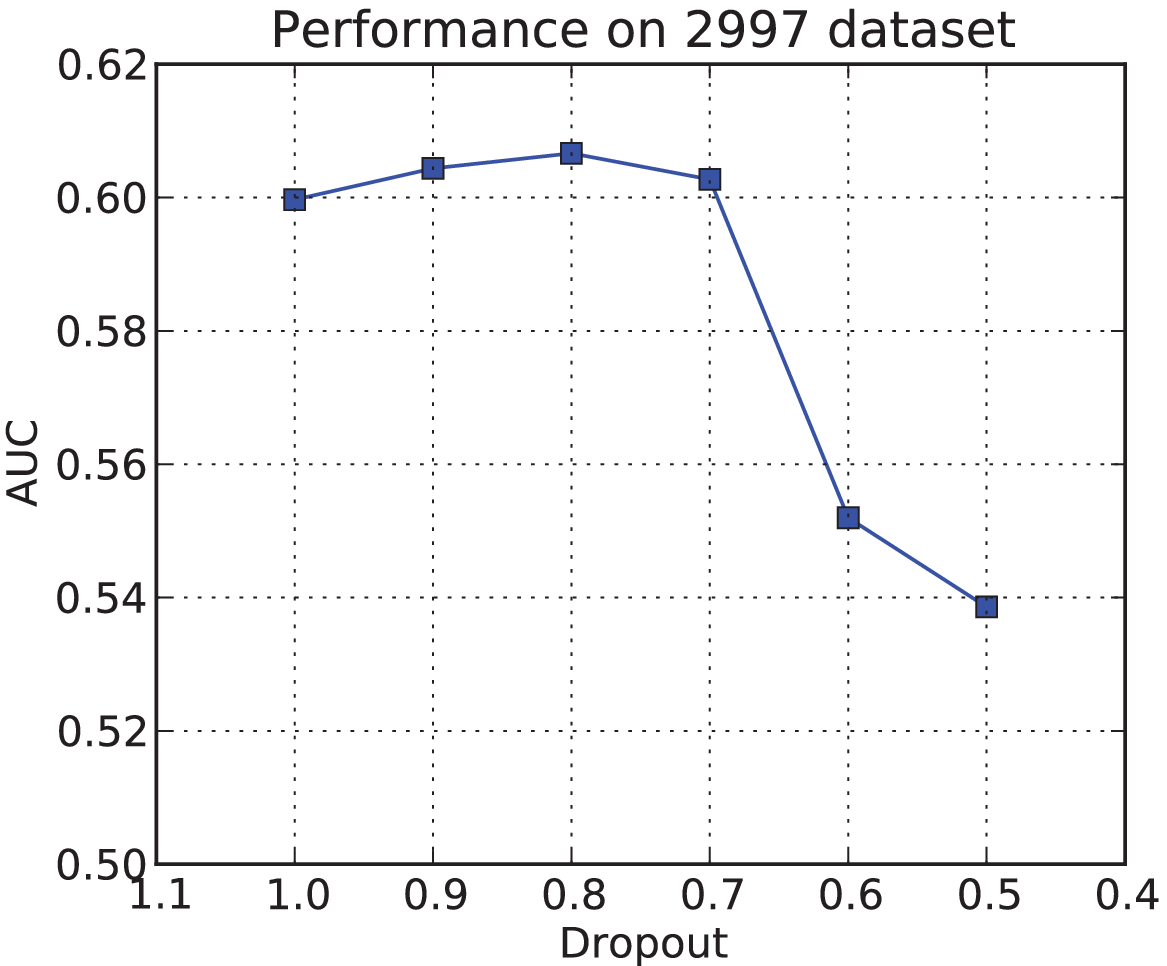} \label{fig:dropout-fmnn}}}%
    \subfigure[\snn~on 2997 dataset]{{\includegraphics[width=0.33\columnwidth]{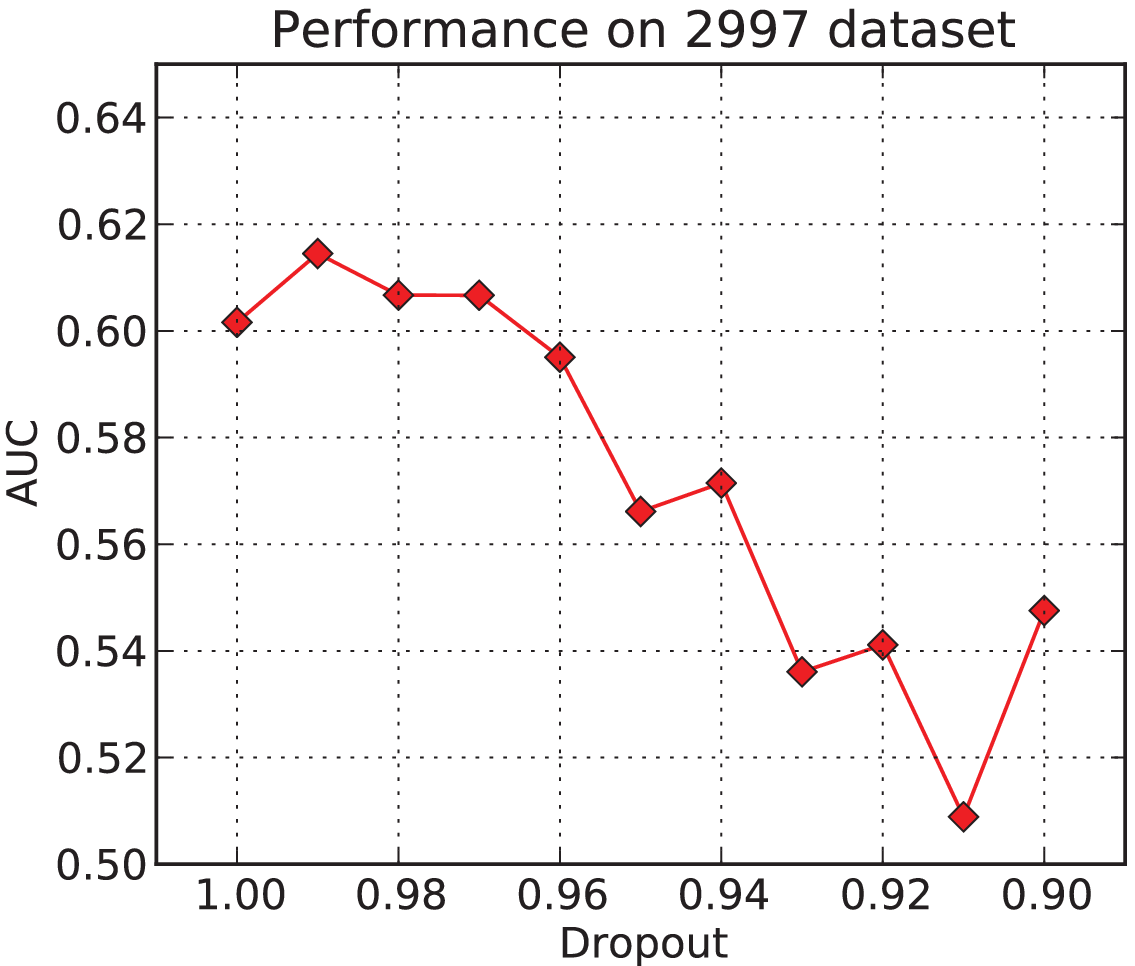} \label{fig:dropout-snn}}}\vspace{-10pt}
    \caption{AUC performance w.r.t difference regularisation settings.}\vspace{-10pt}
    \label{fig:dropoutauc}%
\end{figure}

Figure~\ref{fig:regular} shows the compared AUC performance of \snnrbm~regularised by L2 norm and dropout. It is obvious that dropout outperforms L2 in all compared settings. The reason why dropout is more effective is that when feeding each training case, each hidden unit is stochastically excluded from the network with a probability of dropout rate, i.e., each training case can be regarded as a new model and these models are averaged as a special case of bagging \cite{breiman1996bagging}, which effectively improves the generalisation ability of DNN models.

\subsection{Analysis of Parameters}
As a summary of Sections \ref{sec:struct} and \ref{sec:reg}, for both \fmnn~and \snn, there are two important parameters which should be tuned to make the model more effective: (i) the parameters of layer size decide the architecture of the neural network and (ii) the parameter of dropout rate changes generalisation ability on all datasets compared to neural networks just with L2 regularisation.

Figures~\ref{fig:dropout-fmnn} and \ref{fig:dropout-snn} show how the AUC performance changes with the increasing of dropout in both \fmnn~and \snn. We can find that there is an upward trend of performance in both models at the beginning and then drop sharply with continuous decreasing of dropout rate. The distinction between two models is the different sensitivities of the dropout. From Figure~\ref{fig:dropout-snn}, we can see the model \snn~is sensitive to the dropout rate. This might be caused by the connectivities in the bottom layer. The bottom layer of the \snn~is fully connected with the input vector while the bottom layer for \fmnn~is partially connected and thus the \fmnn~is more robust when some hidden units are dropped out. Furthermore, the sigmoid activation function tend to more effective than the linear activation function in terms of dropout. Therefore, the dropout rates at the best performance of \fmnn~and \snn~are quite different. For \fmnn~the optimal dropout rate is around 0.8 while for \snn~is about 0.99.



\section{Conclusion}\label{sec:con}
In this paper, we investigated the potential of training deep neural networks (DNNs) to predict users' ad click response based on multi-field categorical features. To deal with the computational complexity problem of high-dimensional discrete categorical features, we proposed two DNN models: field-wise feature embedding with supervised factorisation machine pre-training, and fully connected DNN with field-wise sampling-based RBM and DAE unsupervised pre-training. These architectures and pre-training algorithms make our DNNs trained very efficiently. Comprehensive experiments on a public real-world dataset verifies that the proposed DNN models successfully learn the underlying data patterns and provide superior CTR estimation performance than other compared models.
The proposed models are very general and could enable a wide range of future works. For example, the model performance can be improved by momentum methods in that it suffices for handling the curvature problems in DNN training objectives without using complex second-order methods \cite{sutskever2013importance}. In addition, the partial connection in the bottom layer could be extended to higher hidden layers as partial connectivities have many advantages such as lower complexity, higher generalisation ability and more similar to human brain \cite{elizondo1997survey}.

\bibliographystyle{splncs03}
\bibliography{ecir20-zhang}

\end{document}